\documentclass[runningheads]{llncs}
\usepackage[T1]{fontenc}

\usepackage{cite}
\usepackage{amsmath,amssymb,amsfonts}
\usepackage{algorithmic}
\usepackage{graphicx}
\usepackage{textcomp}
\usepackage{xcolor}
\usepackage{tikz}
\usetikzlibrary{calc, arrows}
\usepackage[normalem]{ulem}

\usepackage[hidelinks]{hyperref}
\hypersetup{
    colorlinks=false,
    linkcolor=blue,
    filecolor=magenta,      
    urlcolor=cyan,
    pdftitle={Overleaf Example},
    pdfpagemode=FullScreen,
}

\newcommand{\Real}{\mathbb{R}}
\newcommand{\Natural}{\mathbb{N}}

\newcommand{\sign}{\mathrm{sign}}
\DeclareMathOperator*{\argmin}{arg\,min}

\newtheorem{thm}{Theorem}

\begin{document}

\title{The Boundaries of Verifiable Accuracy, Robustness, and Generalisation in Deep Learning
}

\author{Alexander Bastounis\inst{1}\orcidID{0000-0002-2867-4635} \and
Alexander N. Gorban\inst{1,5}\orcidID{0000-0001-6224-1430} \and
Anders C. Hansen\inst{2} \and Desmond J. Higham\inst{3}\orcidID{0000-0002-6635-3461} \and Danil Prokhorov\inst{4}\orcidID{0000-0002-6208-4233} \and Oliver Sutton\inst{5}\orcidID{0000-0003-0184-4371} \and Ivan Y. Tyukin\inst{5}\orcidID{0000-0002-7359-7966} \and Qinghua Zhou\inst{5}}
\authorrunning{A. Bastounis et al.}
\titlerunning{The Boundaries of Verifiable Accuracy, Robustness, and Generalisation}
\institute{University of Leicester, Leicester LE1 7RH, UK \and
University of Cambridge, Cambridge CB3 0WA, UK \and
University of Edinburgh, Edinburgh EH9 3FD, UK \and Toyota Tech Center, Ann Arbor, USA \and King's College London, London WC2R 2LS, UK}
\maketitle

\begin{abstract}
In this work, we assess the theoretical limitations of determining guaranteed stability and accuracy of neural networks in classification tasks. We consider classical distribution-agnostic framework and algorithms minimising empirical risks and potentially subjected to some weights regularisation. We show that there is a large family of tasks for which computing and verifying ideal stable and accurate neural networks in the above settings is extremely challenging, if at all possible, even when such ideal solutions exist within the given class of neural architectures.


\end{abstract}

\section*{Notation}

{$\Real$ denotes the field of real numbers, $\Real_{\geq 0}=\{x\in\Real| \ x\geq 0\}$, and} $\Real^n$ denotes the $n$-dimensional real vector space, $\Natural$ denotes the set of natural numbers; $({x},{y})=\sum_{k} x_{k} y_{k}$ is the inner product of ${x}$ and ${y}$, and $\|{x}\|=\sqrt{({x},{x})}$ is the standard Euclidean norm  in $\Real^n$; $\mathbb{B}_n$ denotes the unit ball in $\Real^n$ centered at the origin $\mathbb{B}_n=\{{x}\in\Real^n \ | \ {\|{x}\|\leq 1}\}$, $\mathbb{B}_{n} (r,y)$ is the ball in $\mathbb{R}^n$ centred at $y$ with radius $r \geq 0$: $\mathbb{B}_{n} (r,y)=\{x\in\Real^n \ | \ \|{x}-y\| \leq r\}$; $\mathrm{Cb}(\ell, y)$ is the cube in $\Real^n$ centered at $y$ with side-length ${\ell \geq 0}$: $\mathrm{Cb}(\ell, y) = \Big\{x \in\Real^n \ | \ \|x - y \|_{\infty} \leq \frac{\ell}{2}\Big\}$; $\mathbb{S}_{n-1}(r,y)$ is the sphere in $\Real^n$ centred at $y$ with radius $r$: $\mathbb{S}_{n-1}(r,y)=\{{x}\in\Real^n \ | \ {\|{x}-y\|= r}\}$; 
$\sign(\cdot) : \Real \to \Real_{\geq 0}$ denotes the function such that $\sign(s)=1$ for all $s\in\Real_{\geq 0}$ and $\sign(s)=0$ otherwise;  $\mathcal{K}_\theta$ is the class of real-valued functions defined on $\Real$ which are continuous, strictly monotone on $[\theta,\infty)$, and constant on $(-\infty,\theta)$; $\boldsymbol{1}_n$ denotes the vector $(1,\dots,1)\in\Real^n$.

\section{Introduction}

Data-driven AI systems and neural networks in particular have shown tremendous successes across a wide range of applications, including automotive, healthcare, gaming, marketing, and more recently natural language processing. Fuelled by high and growing rates of adoption of the new technology across sectors, robustness and stability are vital characterisations of AI performance. 

The importance  of AI stability and robustness is exemplified by the discovery of adversarial perturbations \cite{szegedy2013intriguing} -- imperceptible changes of input data leading to misclassifications. These perturbations can be  universal \cite{moosavi2017universal} (i.e. triggering misclassifications for many inputs), limited to a single attribute \cite{su2019one}, or masquerading as legitimate inputs \cite{eykholt2018robust}. Sometimes, such AI instabilities can be typical \cite{tyukin2020adversarial}, \cite{shafahi2018adversarial}. Moreover, instabilities can also be induced by perturbations of the AI structure \cite{tyukin2021feasibility}. 

The issue of AI robustness is non-trivial and cannot be considered in isolation from other measures of AI performance: a model returning the same output regardless of the inputs is perfectly robust yet useless. A theoretical framework to approach the problem has recently been proposed in  \cite{bastounis2021mathematics}.
It has been shown in \cite{bastounis2021mathematics} that (i) there is an uncountably large family of distributions such that for an appropriately large data sample drawn from a distribution from this family there is a feed-forward neural network showing excellent performance on this sample, although (ii) this same network becomes inevitably unstable on some subset of the training and validation sets. Moreover, (iii) for the same distribution and the same data, there is a stable network possibly having a different architecture.

Here we show that the stability-accuracy issues have other unexplored dimensions and could be significantly more pronounced than previously thought. Our main result, Theorem~\ref{thm:boundaries_2} shows that there exist large families of well-behaved data distributions for which even networks achieving zero training and validation error may be highly unstable with respect to almost any small perturbation on nearly half of the training or validation data.
Yet, for the same data samples and distributions, there exist stable networks \emph{with the same architecture as the unstable network} which also minimise the loss function.  Strikingly, there exist infinitely many pairs of networks, in which one network is stable and accurate and the other is also accurate but unfortunately unstable, whose weights and biases could be made arbitrarily close to each other. What is even more interesting, all this happens and persists when the values of weights and biases are made small.

This result reveals a fundamental issue at the heart of current data-driven approaches to learning driven by minimising empirical risk functions, even in the presence of weight regularisation, in distribution-agnostic settings. The issues is that such learning algorithms could be structurally incapable of distinguishing between stable and unstable solutions.

The rest of the paper is organised as follows. In Section \ref{sec:preliminaries} we introduce notation and problem setting. In Section \ref{sec:main_results}  we state our main results along with discussion, interpretation, and comparison to the literature.  Section \ref{sec:conclusion} concludes the paper.

\section{Preliminaries, assumptions, and problem settings}\label{sec:preliminaries}

%\subsection{Prior work}

Following \cite{bastounis2021mathematics}, by  $\mathcal{NN}_{\mathbf{N},L}$ we denote the class of neural networks with $L$ layers and dimension $\mathbf{N}=\{N_{L}, N_{L-1}, N_{L-2}, \dots, N_1, N_0=n\}$, where $n$ is the input dimension, and $N_L=1$ is the dimension of the network's output. A neural network with dimension $(\mathbf{N},L)$ is a map
\[
\phi = G^L \sigma G^{L-1}\sigma \cdots \cdots \sigma G^1,
\]
where $\sigma:\Real \rightarrow \Real$ is a coordinate-wise activation function, and 
$
G^{l}: \Real^{N_{l-1}} \rightarrow \Real^{N_l} 
$ 
is an affine map defined by
$G^{l}x = W^{l} x + b^{l}$, where $W^{l}\in\Real^{N_{l}\times N_{l-1}}$, $b^{l}\in\Real^{N_l}$ are the corresponding matrices of weights and biases. By $\Theta(\phi)$ we denote the vector of all weights and biases of the network $\phi$.

In general, the activation functions $\sigma$ do not have to be the same for all components and all layers, although here we will assume (unless stated otherwise) that this is indeed the case. In what follows we will consider feed-forward networks with activation functions in their hidden layers computing mappings from the following broad class:
\begin{equation}\label{eq:activation_functions}
% \begin{split}
% & 
%	g_\theta:\Real\rightarrow\Real, \  g_\theta \in \mathcal{K}_\theta, \ \theta\in \Real.
\sigma=g_\theta,  \  g_\theta \in \mathcal{K}_\theta, \ \theta\in \Real.
% \end{split}
\end{equation}
Popular functions such as ReLU are contained in this class (that is the class of functions which are continuous, strictly monotone on $[\theta,\infty)$ and constant on $(-\infty,\theta)$). 
The condition of strict monotonicity of $g_\theta$ over $[\theta,\infty)$ can be reduced to strict monotonicity over some $[\theta, \theta_1]$, $\theta_1>\theta$, with $g_\theta$ being merely monotone on $[\theta_1,\infty)$. 
This extension won't have any affect on the validity of the theoretical statements below, but will enable the inclusion of leaky ReLU activations (since then activation functions satisfying~\eqref{eq:activation_functions} can be constructed as a difference of a leaky ReLU function and its shifted/translated copy, and the results below therefore still follow) as well as ``sigmoid''-like piecewise linear functions.

We will suppose that all data are drawn from some unknown probability distribution belonging to a family $\mathcal{F}$, and each element $\mathcal{D}\in\mathcal{F}$ of this family is supported on $[-1,1]^n\times\{0,1\}$. 
For any given $\mathcal{D}\in\mathcal{F}$, we will assume that the training and testing algorithms have access to samples $(x^j,\ell^j)$, $j=1,\dots,s+r$, $s,r\in \Natural$, independently drawn from $\mathcal{D}$, and which can be partitioned into training
\[
\mathcal{T}=\{(x^1,\ell^1), \dots, (x^r,\ell^{r}) \}
\]
and validation/testing 
\[
\mathcal{V}=\{(x^{r+1},\ell^{r+1}), \dots, (x^{r+s},\ell^{r+s}) \}
\]
(multi)-sets. Let $M=r+s=|\mathcal{T}\cup\mathcal{V}|$ be the size of the joint training and validation (multi)-set.

Further, we impose a condition that the data distribution is sufficiently regular and does not possess hidden instabilities and undesirable accumulation points which could otherwise trivialise our statements and results. 
In particular, for $\delta\in (0,2\sqrt{n}]$ we will only consider those distributions $\mathcal{D}_\delta\in\mathcal{F}$ which satisfy:
\begin{equation}\label{eq:separability}
\text{If } (x,\ell_x),(y,\ell_y)\sim \mathcal{D}_\delta \text{ with } \ell_x\neq\ell_y, \text{ then, with probability 1,} \ \|x-y\|\geq \delta.
\end{equation}

Finally, we introduce the family of loss functions
\begin{equation}\label{eq:loss_function}
\begin{split}
\mathcal{CF}_{\mathrm{loc}}=&\{\mathcal{R}: \ \Real\times\Real\rightarrow\Real_{\geq 0}\cup \{\infty\} \ | \ \mathcal{R}(v,w)=0 \ \iff \ v=w \}
\end{split}
\end{equation}
which will be used to define the corresponding empirical loss functions for the model outputs $h:\Real^n\rightarrow\{0,1\}$ on samples $\mathcal{S}\sim \mathcal{D}_\delta$ drawn from $\mathcal{D}_\delta$
\begin{equation}\label{eq:loss_functional}
\mathcal{L}(\mathcal{S},h)=\sum_{(x^i,\ell^i)\in\mathcal{S}} \mathcal{R}(h(x^i),\ell^i).
\end{equation}
The subscript ``$\mathrm{loc}$'' in (\ref{eq:loss_function}) emphasises that the loss functions $\mathcal{R}$ are evaluated on single data points and in this sense are  ``local''. It provides an explicit connection with the classical literature involving empirical risk minimisation, allowing us to exploit the conventional interpretation of the generalisation error as a deviation of the empirical risk from the expected value of the loss over the distribution generating the data. 

\section{Main results}\label{sec:main_results}

Having introduced all relevant notation, are now ready to state the main result of the contribution.

\begin{thm}[Inevitability, typicality and undetectability of instability]\label{thm:boundaries_2} 
Consider the class of networks with architecture 
\[
{\bf{N}}=(N_L=1,N_{L-1},\dots,N_{1},N_0=n), \  \ L\geq 2, \ n\geq 2,
\]
where $N_1\geq 2n$ and $N_2,\dots, N_{L-1}\geq 1$, and activation functions $g_\theta$ in layers $1,\dots,L-1$ satisfying conditions (\ref{eq:activation_functions}), and the $\sign(\cdot)$ activation function in layer $L$. 

Let $\varepsilon\in(0,\sqrt{n}-1)$ and fix $0<\delta\leq \varepsilon/\sqrt{n}$.
Then, there is an uncountably large family of distributions $\mathcal{D}_\delta \in \mathcal{F}$ satisfying~\eqref{eq:separability} such that for any $\mathcal{D}_\delta \in \mathcal{F}$, any training and validation data $\mathcal{T}$, $\mathcal{V}$ drawn independently from $\mathcal{D}_\delta$, and every $\mathcal{R}\in\mathcal{CF}_{\mathrm{loc}}$, with probability 1:

\begin{itemize}
\item[(i)] There exists a network which correctly classifies the training data $\mathcal{T}$ and generalises to the test data $\mathcal{V}$, satisfying
\[
f \in \argmin_{\varphi\in \mathcal{NN}_{{\bf{N}},L}} \mathcal{L}(\mathcal{T}\cup\mathcal{V},\varphi)
\]
with
$
\mathcal{L}(\mathcal{T}\cup\mathcal{V},f) = 0.
$

Yet, for any $q\in(0,1/2)$, with probability greater than or equal to
\[
1-\exp(-2q^2 M)
\]
there exists a multi-set $\mathcal{U}\subset \mathcal{T}\cup\mathcal{V}$ of cardinality at least  $\lfloor(1/2-q) M\rfloor$ on which $f$ is unstable in the sense that for any $(x,\ell)\in\mathcal{U}$ and any $\alpha\in(0,\varepsilon/2)$, there exists a perturbation $\zeta \in \Real^n$ with $\|\zeta\| \leq \alpha/\sqrt{n}$ and
\begin{equation}\label{eq:instability}
	|f(x)-f(x+\zeta)|=1.
\end{equation}
Moreover, such destabilising perturbations are \emph{typical} in the sense that if vectors $\zeta$ are sampled from the equidistribution in $\mathbb{B}_n(\alpha/\sqrt{n}, 0)$, then for $(x,\ell) \in \mathcal{U}$, the probability that~\eqref{eq:instability} is satisfied is at least
\[
1-\frac{1}{2^n}.
\]
Furthermore, there exist \emph{universal} destabilising perturbations, in the sense that a single perturbation $\zeta$ drawn from the equidistribution in $\mathbb{B}_n(\alpha/\sqrt{n}, 0)$ destabilises $m \leq |\mathcal{U}|$ points from the set $\mathcal{U}$ with probability at least
\[
1-\frac{m}{2^n}.
\]

\item[(ii)] At the same time, for the same distribution $\mathcal{D}_\delta$ there is a robust network with the same architecture as $f$, satisfying 
\[
\tilde{f}\in\argmin_{\varphi\in \mathcal{NN}_{{\bf{N}},L}} \mathcal{L}(\mathcal{T}\cup\mathcal{V},\varphi)
\]
with
$
\mathcal{L}(\mathcal{T}\cup\mathcal{V},\tilde{f}) = 0,
$
which is robust in the sense that for all $(x,\ell)\in \mathcal{T}\cup\mathcal{V}$
\[
% \begin{split}
	\tilde{f}(x)=\tilde{f}(x+\zeta)
% \end{split}
\]
for any $\zeta \in \Real^n$ with $\|\zeta\| \leq \alpha/\sqrt{n}$, even when $|\mathcal{T} \cup \mathcal{V}| = \infty$.

Moreover, there exist pairs of unstable and robust networks, $f_\lambda, \tilde{f}_\lambda$ and $f_\Lambda, \tilde{f}_\Lambda$, satisfying the statements above such that the maximum absolute difference between their weights and biases is either arbitrarily small or arbitrarily large. That is, for any  $\lambda>0, \Lambda>0$:
\[
\|\Theta(f_\lambda)-\Theta(\tilde{f}_{\lambda})\|_{\infty}<\lambda, \ \|\Theta(f_\Lambda)-\Theta(\tilde{f}_{\Lambda})\|_{\infty}>\Lambda.
\]

\item[(iii)] However, for the above robust solution $\tilde{f}$, 
\begin{itemize} 
\item[a)] there exists an uncountably large family of distributions $\tilde{D}_{\delta}\in\mathcal{F}$ on which $\tilde{f}$ correctly classifies both the training and test data, yet fails in the same way as stated in (i).
\item[b)] there exists an uncountably large family of distributions $\hat{D}_{\delta}\in\mathcal{F}$ such that
the map $\tilde{f}$ is robust on $\mathcal{T}\cup\mathcal{V}$ (with respect to perturbations $\zeta$ with $\|\zeta\|\leq \alpha/\sqrt{n}$, $\alpha\in(0,\varepsilon/2)$) with probability
\[
\left(1-\frac{k}{2^{n+1}}\right)^{M}
\]
but is unstable to arbitrarily small perturbations on a future sample with probability $k/2^{n+1}$.
\end{itemize}
\end{itemize}
\end{thm}

The proof of the theorem is provided in the Appendix.

\subsection{Interpretation of results}

According to statement (i) of Theorem \ref{thm:boundaries_2}, not only are instabilities to be expected, but they can also be remarkably widespread: for sufficiently large data sets they may occur, with high probability, for nearly half of all  data.

Statement (ii) of Theorem \ref{thm:boundaries_2} confirms that a stable solution exists {\it within precisely the same class of network architectures}, although it is difficult to compute it by using only the loss functional $\mathcal{L}$ as a measure of quality.
This shows that the architecture isn't necessarily the source of the instability.
Moreover,  a robust solution may be found in an arbitrarily small neighborhood of the specific non-robust one in the space of network weights and biases.
As the construction in the proof shows, using networks with small Lipshitz constants can, counterintuitively, make the problem worse.

The robust solution, in turn, can also be unstable, as follows from statement (iii), part (a). This is reminiscent of a ``no free lunch'' principle for robust and accurate learning, although with a subtle distinction. 
In fact, as part b) of the statement states, there are solutions which may appear to be certifiably robust (and one can indeed certify the model on the training and validation sets), although there is no guarantee whatsoever that the certificate remains valid for future samples.
To minimise the risks, one needs to certify the model on data sets which are exponentially large in $n$.
This is particularly relevant for safety-critical settings, where the risk of failure must be calculated and bounded in advance.

Finally, we note that the instabilities considered in Theorem \ref{thm:boundaries_2} become particularly pronounced for networks with sufficiently high input dimension $n$ (see statement (iii) of the theorem). 
Moreover, statement (ii) shows that the fraction of perturbations around unstable points $x$ in the sample which alter the network's response approaches $1$ as $n$ grows.  
These high-dimensional effects may still be observed in networks with arbitrarily low input dimensions if such networks realise appropriate auxiliary space-filling mappings in relevant layers.
The technical point that the statement of Theorem~\ref{thm:boundaries_2} holds with probability one is due to the fact that the proof constructs data distributions which assign probability zero to certain sets, so there may exist training samples with probability zero for which the construction does not apply.

\subsection{Discussion}

\subsubsection{Instabilities and regularisation}

The construction we used in the proof of Theorem \ref{thm:boundaries_2} reveals that the instability  discussed in statements (i) and (ii) of the theorem is inherent to the very definition of the binary classification problem and may not be addressed by regularisation approaches constraining norms of network's parameters and Lipschitz constants of non-threshold layers. 

Indeed, consider just the first two layers of the network $f$ constructed in the proof of the theorem, remove the $\sign(\cdot)$ activation function, and introduce an arbitrarily small positive factor $\beta$ (cf. (\ref{eq:network_core})):
\begin{equation}\label{eq:network_core:Lipschitz}
\begin{split}
f_{\mathrm{reg}}(x)=&\sum_{i=1}^n g_\theta(\theta) - g_\theta (\beta((x,e_i)  -  1/\sqrt{n}) + \theta) \\
& +  \sum_{i=1}^n g_\theta(\theta) - g_\theta(\beta(- (x,e_i) -1/\sqrt{n}) + \theta).
\end{split}
\end{equation}
If the functions $g_\theta$ are Lipschitz then the Lipschitz constant of the function $f_{\mathrm{reg}}$ can be made arbitrarily small by setting $\beta$ to some sufficiently small value. At the same time, the values of $\sign f_{\mathrm{reg}}(x)$ and $f(x)$ coincide. This implies that regardless of how well-behaved the function $f_{\mathrm{reg}}$ in (\ref{eq:network_core:Lipschitz}) is, forced classification achieved either by the application of the $\sign$ function or, alternatively, through thresholding or softmax, brings instabilities. 

In this respect, network regularisation by pruning, restricting norms of the network's weights, and forcing the network's Lipschitz constant to stay small do not always warrant robustness. 
Similarly, requesting that there is some non-zero margin separating the classes does not address or alleviate the problem either. 
The instability occurs due to the fact that the algorithm is required to produce a decision boundary, but is unaware that the data is placed directly on this boundary.

\subsubsection{Adversarial training}

A potential way to overcome the instabilities formalised in statement (i) of Theorem \ref{thm:boundaries_2} is to invoke a type of training capable of assessing that instabilities (\ref{eq:instability}) do not occur. 
Adversarial training and data augmentation, whereby each data sample produces a set of points corresponding to perturbed data is an example of an approach which can potentially address the problem. 
The approach is not without its own challenges as one needs to ensure that all points in the sets $\mathbb{B}_n(\alpha/n,x)$, $\alpha\in(0,\varepsilon/2)$ are checked. The latter task can be computationally and numerically overwhelming for large $n$.

\subsubsection{Dark data}

The final and perhaps the most interesting point in relation to the problem of verifiability is statement (iii), which can be related to challenge of the ``dark data'' -- the data which exists but to which we don't have access \cite{schembera2020dark} or, more generally, the missing data and the data which we don't have \cite{hand2020dark}. 
As the theorem states, high-dimensional distributions could be a very real source of such dark data, potentially leading to instabilities or non-verifiability.   

\section{Conclusion}\label{sec:conclusion}

Deep learning networks and models have convincingly shown ample capabilities in many practical tasks. 
When properly engineered, these models stunningly outperform shallower architectures (see e.g. \cite{e24111635}, \cite{yarotsky2017error} for examples and precise statements). 
Moreover, recent breakthroughs such as the emergence of ChatGPT show exceptional power these models may bring. 
These models operate in high-dimensional spaces and process and execute decisions on genuinely high-dimensional data.

At the same time, and despite these remarkable achievements, the application of these highly expressive and capable models requires special care and understanding of their fundamental limitations.

Our work, by building on \cite{bastounis2021mathematics}, reveals a new set of limitations which are particularly inherent to high-dimensional data. 
These limitations constitute the presence of nested uncountably large families of exceptions on which even moderately-sized networks may and likely will fail. 
The results also show that it may be computationally hard to verify both robustness and accuracy of models within classical distribution-agnostic learning frameworks based solely on the notions of risk and empirical risk minimisation. 
All these call for the need to rethink standard distribution-agnostic learning frameworks and introduce more appropriate models of reality into the mathematical setting of statistical learning. 
 
The results, by showing fundamental difficulties with guaranteeing simultaneous stability, accuracy, and verifiability, highlight the importance of mathematical theory and methods for the continuous correction of AI models \cite{gorban2016blessing}, \cite{gorban2017stochastic}, \cite{gorban2021high}. 
 
At present, the results do not include networks with classical sigmoidal activation functions. Detailed analysis of these types of networks will be the topic of our future work.

\subsubsection{Acknowledgements} This work is supported by the UKRI, EPSRC [UKRI Turing AI Fellowship ARaISE EP/V025295/2 and UKRI Trustworthy Autonomous Systems Node in Verifiability EP/V026801/2 to I.Y.T.,  EP/V025295/2 to O.S., A.N.G., and Q.Z.,  EP/V046527/1 and EP/P020720/1 to D.J.H,  EP/V046527/1 to A.B.]. 

\bibliographystyle{splncs04}
\bibliography{limitations_refs}

\section*{Appendix}

\subsection{Proof of Theorem \ref{thm:boundaries_2}}

\subsubsection{Proof of statement (i) of the theorem}

The proof consists of three parts. 
The first part introduces a family of distributions satisfying the separability requirement \eqref{eq:separability} and shows relevant statistical properties of samples drawn from these distributions. 
The second part presents the construction of a suitable neural network minimising the empirical loss function $\mathcal{L}$ for any loss function $\mathcal{R}\in\mathcal{CF}_{\mathrm{loc}}$ which successfully generalises beyond training (and test/validation) data. 
The final part shows that, with high probability, this network is unstable on nearly half of the data (for $s+r$ reasonably large).

{\it Proof of statement (i), part 1}. 
Consider the $n$-dimensional hyper cube $\mathrm{Cb}(2, 0)$ $=$ $[-1,1]^n$.
Within this cube, we may inscribe the unit ball $\mathbb{B}_{n}$ (the surface of which touches the surface of the outer cube at the centre of each face), and within this ball we may, in turn, inscribe the inner cube $\mathrm{Cb}(2/\sqrt{n}, 0)$ each vertex of which touches the surface of the ball and whose faces are parallel to the faces of the cube $\mathrm{Cb}(2, 0)$.
For any $\varepsilon\in(0,\sqrt{n}-1)$, the cube 
$
\mathrm{Cb}(\frac{2}{\sqrt{n}}(1+\varepsilon), 0)
$
may be shown to satisfy $\mathrm{Cb}(2/\sqrt{n}, 0) \subset \mathrm{Cb}(\frac{2}{\sqrt{n}}(1+\varepsilon), 0) \subset \mathrm{Cb}(2, 0)$.

Let
$
V = \{v_i\}_{i=1}^{2^n}
$
denote the set of vertices of $\mathrm{Cb}(2/\sqrt{n}, 0)$ with an arbitrary but fixed ordering, and note that each $v_i$ may be expressed as $\frac{1}{\sqrt{n}}(q_1,\dots,q_n)$ with each component $q_k\in\{-1,1\}$.
The choice of $\varepsilon$ ensures that the set
\[
\mathcal{J}_0=\Big\{x\in\mathbb{S}_{n-1}(1,0) \ | \ x\notin \mathrm{Cb}\Big(\frac{2}{\sqrt{n}}(1+\varepsilon), 0\Big) \Big\}.
\]
is non-empty and that
$ \min_{x\in\mathcal{J}_0, \ y\in V} \|x-y\| > \frac{\varepsilon}{\sqrt{n}}$.

Consider a family of distributions $\mathcal{F}_1\subset \mathcal{F}$ which are supported on $\mathbb{S}_{n-1}(1,0)\times\{0,1\}$, with the $\sigma$-algebra $\Sigma_{\mathbb{S}}\times\{0\}\cup \Sigma_{\mathbb{S}}\times\{1\}$, where $\Sigma_{\mathbb{S}}$ is the standard $\sigma$-algebra on the sphere $\mathbb{S}_{n-1}$ with the topology induced by the arclength metric. 

We construct $\mathcal{F}_1$ as those distributions $\mathcal{D}_\delta \in \mathcal{F}$ such that
\begin{equation}\label{eq:property_distribution:1}
\begin{split}
& P_{\mathcal{D}_\delta}(x,\ell)=0
% \\  & 
\mbox{ for} \ x\in \mathrm{Cb}\left(\frac{2}{\sqrt{n}}(1+\varepsilon), 0\right)\setminus  V, \ \mbox{and any } \ell,
\end{split}
\end{equation}
with
\begin{equation}\label{eq:property_distribution:2}
P_{\mathcal{D}_\delta}(x,\ell)= 
% \left\{\begin{array}{ll}
% 						 \frac{1}{2^{n+1}}, &  x\in V, \ \ell=1\\
% 						 0, & x\in V, \ \ell=0,
% 						 \end{array}\right.
	\begin{cases}
		% 0 & \text{ for } x\in \mathrm{Cb}\left(\frac{2}{\sqrt{n}}(1+\varepsilon), 0\right)\setminus  V
		% \\
		\frac{1}{2^{n+1}} & \text{ for }  x\in V, \ \ell=1
		\\
		0, & \text{ for } x\in V, \ \ell=0
	\end{cases}
\end{equation}
and
\begin{equation}\label{eq:property_distribution:3}
P_{\mathcal{D}_\delta}(\mathcal{J}_0,\ell) = 
	\begin{cases}
		0& \text{ for } \ell=1,\\
		\frac{1}{2}& \text{ for } \ell=0.
	\end{cases}
\end{equation}
The existence of an uncountable family of distributions $\mathcal{D}_\delta$ satisfying \eqref{eq:property_distribution:1}--\eqref{eq:property_distribution:3} 
% exist, with
% \[
% \delta>\frac{\varepsilon}{\sqrt{n}}.
% \]
% Moreover, there are uncountably many of such distributions due to 
is ensured by the flexibility of~\eqref{eq:property_distribution:3} and the fact that $\mathcal{J}_0$ contains more than a single point (consider e.g. the family of all delta-functions supported on $\mathcal{J}_0$ and scaled by $1/2$).
This construction moreover ensures that any $\mathcal{D}_\delta \in \mathcal{F}_1$ also satisfies the separation property~\eqref{eq:separability} with $\delta \leq \frac{\varepsilon}{\sqrt{n}}$.

Let
$
\mathcal{M}=\mathcal{T}\cup\mathcal{V} = \{(x_k, \ell_k)\}_{k=1}^{M},
$
denote the (multi-)set corresponding to the union of the training and validation sets independently sampled from $\mathcal{D}_{\delta}$, where
$
M=s+r=|\mathcal{M}|.
$
% Recall that the elements of $\mathcal{M}$ are all independent samples from the same distribution $\mathcal{D}_\delta$. Introduce a function
Let $z:\Real^n\times\{0,1\} \rightarrow \{0,1\}$ be the trivial function
mapping a sample $(x,\ell)$ from $\mathcal{D}_\delta$ into $\{0,1\}$ by
$
z(x,\ell)=\ell.
$
This function defines new random variables $Z_k = z(x_k, \ell_k) \in [0,1]$ for $k = 1, \dots, M$, with expectation
$
E(Z_k)=\frac{1}{2}.
$

The Hoeffding inequality ensures that
\[
P\left( \frac{1}{2} - \frac{1}{M}\sum Z_k > q \right) \leq \exp\left(-2 q^2 M\right),
\]
and hence, with probability greater than or equal to
\begin{equation}\label{eq:probability_fail_1}
1 - \exp\left(-2 q^2 M\right),
\end{equation}
the number of data points $(x,\ell)$ with $\ell=1$ in the sample $\mathcal{M}$ is at least
\begin{equation}\label{eq:proportion_points}
\lfloor \left(\frac{1}{2} - q\right) M \rfloor.
\end{equation}

{\it Proof of statement (i), part 2}. Let $\{e_1,\dots,e_n\}$ be the standard basis in $\Real^n$.  Consider the following set of $2n$ inequalities:
\begin{equation}\label{eq:basic_inequalities}
(x,e_i) \leq \frac{1}{\sqrt{n}}, \ 
(x,e_i) \geq -\frac{1}{\sqrt{n}}, \text{ for } i=1,\dots,n.
\end{equation}
Any function defined on $[-1/\sqrt{n},1/\sqrt{n}]^n$ (or which contains  $[-1/\sqrt{n},1/\sqrt{n}]^n$ in the domain of its definition) and which returns $1$ for $x$ satisfying (\ref{eq:basic_inequalities}) and $0$ otherwise, minimises the loss $\mathcal{L}$ on $\mathcal{T}$. It also generalises perfectly well on any $\mathcal{V}$. Hence a network implementing such a function shares the same properties.

Pick a function $g_\theta\in\mathcal{K}_\theta$ and consider
\[
g_\theta((x,e_i)  -  1/\sqrt{n} + \theta ), \ g_\theta(-(x,e_i)  +  1/\sqrt{n} + \theta), \ i=1,\dots,n.
\]
It is clear that 
$g_\theta(\theta) - g_\theta ((x,e_i)  -  1/\sqrt{n} + \theta) = 0$
for $(x,e_i) \leq 1/\sqrt{n}$, and 
$g_\theta(\theta) - g_\theta ((x,e_i)  -  1/\sqrt{n} + \theta) < 0$
for $(x,e_i) > 1/\sqrt{n}$. Similarly 
$g_\theta(\theta) - g_\theta(- (x,e_i) -1/\sqrt{n} + \theta) = 0$
for $(x,e_i) \geq - 1/\sqrt{n}$, and 
$g_\theta(\theta) - g_\theta (-(x,e_i)  -  1/\sqrt{n} + \theta) < 0$  
for $(x,e_i) < - 1/\sqrt{n}$.
Hence, the function $f$ given by
\begin{equation}\label{eq:network_core}
\begin{split}
f(x)=&\sign\left(\sum_{i=1}^n g_\theta(\theta) - g_\theta ((x,e_i)  -  1/\sqrt{n} + \theta)\right. \\
& + \left. \sum_{i=1}^n g_\theta(\theta) - g_\theta(- (x,e_i) -1/\sqrt{n} + \theta)\right)
\end{split}
\end{equation}
is exactly $1$ only when all inequalities (\ref{eq:basic_inequalities}) hold true, and is zero otherwise. 
We may therefore conclude that
\[
f\in \arg \min_{\varphi\in \mathcal{NN}_{{\bf{N}},L}} \mathcal{L}(\mathcal{T}\cup\mathcal{V},\varphi).
\]

Observe now that (\ref{eq:network_core}) is a two-layer neural network with $2n$ neurons in the hidden layer and a  threshold output. This core network can be extended to any larger size without changing the map $f$ by  propagating the argument of $\sign(\cdot)$  in (\ref{eq:network_core}) to the next layers and appending the width as appropriate.

{\it Proof of statement (i), part 3}. Let us now show that the map (\ref{eq:network_core}) becomes unstable for an appropriately-sized set $\mathcal{M}$. Suppose that there are $\lfloor(1/2-q)M\rfloor$ data points on which $f(x)=1$, and by construction each is a vertex of $\mathrm{Cb}(2/\sqrt{n}, 0)$.
According to (\ref{eq:probability_fail_1}), (\ref{eq:proportion_points}), the probability of this event is not zero.
Let $x$ be one such point and let $\zeta$ be a perturbation sampled from an equidistribution in the ball $\mathbb{B}_n(\alpha/\sqrt{n}, 0)$ for some $\alpha\in (0,\varepsilon/2)$.
Then, with probability $1 - \frac{1}{2^n}$, the perturbation $\zeta$ is such that
$|f(x + \zeta) - f(x)| = 1$,
since this is true for any $\zeta$ such that $x + \zeta \notin \mathcal{I} = \mathrm{Cb}(2/\sqrt{n}, 0) \cap \mathbb{B}_n(\alpha / \sqrt{n}, x)$, and the set $\mathcal{I}$ is uniquely defined by the signs of exactly $n$ linear inequalities which slice the ball into $2^n$ pieces of equal volume and so has probability $\frac{1}{2^n}$.

Finally, note that if there are at least $m$ points $(u^1,\ell^1),\dots,(u^m,\ell^m)$ in the set $\mathcal{U}$ then the probability that all $u^i+\zeta$, $i=1,\dots,m$ are outside of the corresponding intersections follows from the union bound, which completes the argument.

\subsubsection{Proof of statement (ii) of the theorem}

The argument used in the proof of statement (i), part 2, implies that there exists a network $\tilde{f}\in\mathcal{NN}_{{\bf{N}},L}$ such that $\tilde{f}(x)$ takes value 1 when the inequalities
\begin{equation}\label{eq:basic_inequalities_robust}
 (x,e_i) \leq \frac{1}{\sqrt{n}} \Big( 1+\frac{\varepsilon}{2} \Big),  \ 
 (x,e_i) \geq - \frac{1}{\sqrt{n}} \Big( 1+\frac{\varepsilon}{2} \Big), \text{ for } i=1,\dots,n.
\end{equation}
are satisfied, and zero otherwise.
This network also minimises $\mathcal{L}$ and generalises beyond the training and validation data.

However, since for any $\alpha\in(0,\varepsilon/2)$ the function $\tilde{f}$ is constant within a ball of radius $\alpha/\sqrt{n}$ around any data point $x \in \mathcal{T} \cup \mathcal{V}$, we can conclude that $\tilde{f}$ is insusceptible to the instabilities affecting $f$.

To show that there exists a pair of unstable and stable networks, $f$ and $\tilde{f}$ (the network $\tilde f$ is stable with respect to perturbations $\zeta: \ \|\zeta\|\leq\alpha/\sqrt{n}$), consider systems of inequalities (\ref{eq:basic_inequalities}), (\ref{eq:basic_inequalities_robust}) with both sides multiplied by a positive constant $\kappa>0$. Clearly, and regardless of the multiplication by $\kappa$, these systems of inequalities define the cubes $\mathrm{Cb}(2\sqrt{n},0)$ and $\mathrm{Cb}(2\sqrt{n}(1+\varepsilon/2),0)$, respectively. Then
\begin{equation}\label{eq:network_core:2:unstable}
\begin{split}
f(x)=&\sign\left(\sum_{i=1}^n g_\theta(\theta) - g_\theta (\kappa((x,e_i)  -  1/\sqrt{n}) + \theta)\right. \\
& + \left. \sum_{i=1}^n g_\theta(\theta) - g_\theta(\kappa(- (x,e_i) -1/\sqrt{n}) + \theta)\right)
\end{split}
\end{equation}
encodes the unstable network, and 
\begin{equation}\label{eq:network_core:2:stable}
\begin{split}
\tilde f(x)=&\sign\left(\sum_{i=1}^n g_\theta(\theta) - g_\theta (\kappa((x,e_i)  -  (1+\varepsilon/2)/\sqrt{n}) + \theta)\right. \\
& + \left. \sum_{i=1}^n g_\theta(\theta) - g_\theta(\kappa(- (x,e_i) -(1+\varepsilon/2)/\sqrt{n}) + \theta)\right)
\end{split}
\end{equation}
encodes the stable one. These networks share the same weights but their biases differ in absolute value by $\kappa \varepsilon/(2\sqrt{n})$. Given that $\kappa$ can be chosen arbitrarily small or arbitrarily large, the statement now follows.

\subsubsection{Proof of statement (iii) of the theorem}

Part a) of statement (iii) can be demonstrated following the same argument used to prove of statement (i) by replacing the cube $\mathrm{Cb}(2/\sqrt{n}, 0)$ with $\mathrm{Cb}(2/\sqrt{n}(1+\varepsilon/2), 0)$.

Part b) follows by considering a slightly modified family of distributions $\hat{\mathcal{D}}_\delta$ in which the set $V$ that was used to define $\mathcal{D}_\delta$ is replaced with the set
\[
V=\{v_i \ | \ i=1,\dots,2^n - k \} \cup \hat{V},
\]
where
\[
	\hat{V}=\{v_i (1 + \varepsilon/2) \ | \ i=2^n-k+1,\dots, 2^n\}.
\]
The probability that a single point from the set $\hat{V}$ is sampled in a single independent draw from $\hat{\mathcal{D}}_\delta$ is $k/2^{n+1}$ (this follows from the union bound).  Since all points in the multi-set $\mathcal{V}\cup\mathcal{T}$ are drawn independently, the probability that none of these points are present in $\mathcal{M}$ is $(1-k/2^{n+1})^{M}$.
\qed

\end{document}